\documentclass[10pt,twocolumn,letterpaper]{article}

\usepackage{iccv}     
\usepackage{adjustbox} 
\usepackage{booktabs}
\usepackage{graphicx} 
\usepackage{comment}
\usepackage{multirow}
\usepackage{graphicx}
\usepackage{caption}
\usepackage{marvosym} 
\usepackage{array}
\usepackage{colortbl} 
\usepackage{adjustbox} 
\usepackage[table]{xcolor}
\usepackage{xcolor}

\usepackage{booktabs}
\usepackage{arydshln}
\usepackage{tabularx}
\usepackage{amssymb}

\definecolor{iccvblue}{rgb}{0.21,0.49,0.74}
\usepackage[pagebackref,breaklinks,colorlinks,allcolors=iccvblue]{hyperref}

\def\paperID{3dqXxlHS6k} 
\def\confName{ICCV}
\def\confYear{2025}

\title{FinCap: Topic-Aligned Captions for Short-Form Financial YouTube Videos}

\author{
Siddhant Sukhani\textsuperscript{1,2,*\,\Letter}\quad
Yash Bhardwaj\textsuperscript{2*}\quad
Riya Bhadani\textsuperscript{2}\quad
Veer Kejriwal\textsuperscript{2}\quad \\
Michael Galarnyk\textsuperscript{2}\quad
Sudheer Chava\textsuperscript{2}\\
\textsuperscript{1}\,Stanford University \\
\textsuperscript{2}\,Georgia Institute of Technology\\
{\Letter\;\;Corresponding author: \href{mailto:sukhani@stanford.edu}{sukhani@stanford.edu}}
}

\begin{document}
\maketitle
\footnotetext{*\;These authors contributed equally.}
\begin{abstract}
We evaluate multimodal large language models (MLLMs) for topic-aligned captioning in financial short-form videos (SVs) by testing joint reasoning over transcripts (T), audio (A), and video (V). Using 624 annotated YouTube SVs, we assess all seven modality combinations (T, A, V, TA, TV, AV, TAV) across five topics: main recommendation, sentiment analysis, video purpose, visual analysis, and financial entity recognition. Video alone performs strongly on four of five topics, underscoring its value for capturing visual context and effective cues such as emotions, gestures, and body language. Selective pairs such as TV or AV often surpass TAV, implying that too many modalities may introduce noise. These results establish the first baselines for financial short-form video captioning and illustrate the potential and challenges of grounding complex visual cues in this domain. All code and data can be found on our Github\footnote{\href{https://github.com/gtfintechlab/FinCap}{https://github.com/gtfintechlab/FinCap}} under the CC-BY-NC-SA 4.0 license.
\end{abstract}
    
\section{Introduction}
\label{sec:intro}
\begin{figure}[htbp]
\centering
\includegraphics[width=\columnwidth]{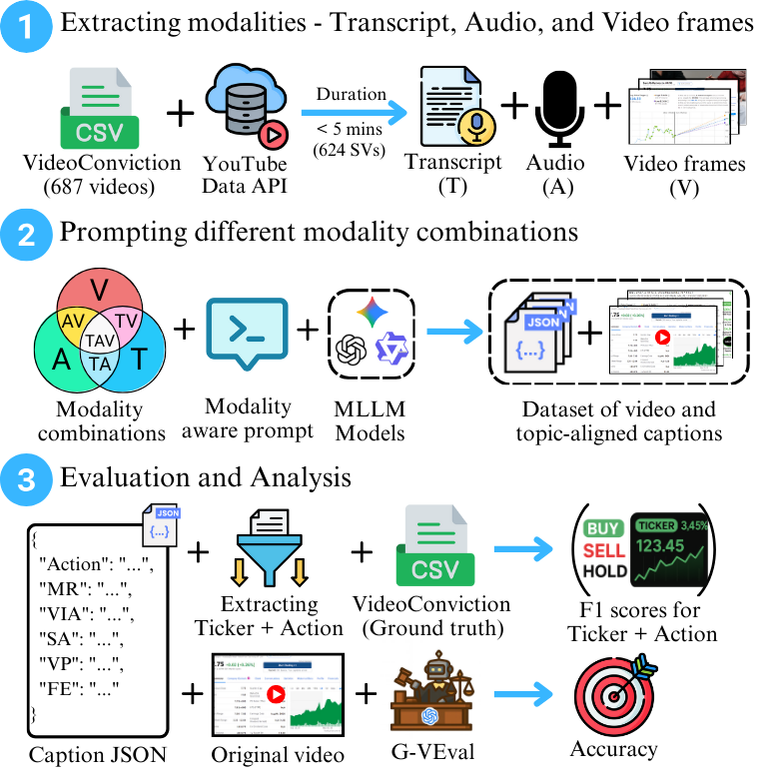}
\caption{End-to-end pipeline: (1) Extract text, audio, and video frames from VideoConviction \cite{galarnyk2025videoconviction}, leading to 624 SVs; (2) prompt MLLMs with all non-empty modality pairs to generate topic-aligned captions; (3) evaluate captions with G-VEval for accuracy and weighted F1 score on ticker–action pairs.}

\label{fig:method}
\end{figure}
Short-form videos (SVs) have become a dominant medium for information exchange, condensing dense multimodal signals into seconds \cite{ge2025archunyuanvideo7bstructuredvideocomprehension,Zhang2025Engagement}. Platforms like YouTube host billions of videos \cite{YouTube2025TwentyBillion}, where rapid visual transitions, layered audio, and textual overlays are tightly interwoven \cite{smith2022rise, lee2023youtube}. Capturing structured meaning from such content goes well beyond generic video captioning, demanding models that integrate visual, auditory, and textual cues with domain-specific reasoning. While multimodal large language models (MLLMs) have advanced video understanding \cite{venugopalan2015seq, xu2016msrvtt, tang2024enhancing, ventura2025chapterllamaefficientchapteringhourlong, yang2023vidchapters7mvideochaptersscale}, SVs remain largely unexplored \cite{yu2024prompting}. Financial SVs present unique challenges: (i) overlapping on-screen elements such as charts, stock tickers (e.g., Microsoft’s ticker is MSFT), logos, and annotations that demand robust object, scene, and multi-concept recognition; (ii) dense financial terminology and abbreviations that require precise grounding; and (iii) cues dispersed across non-contiguous segments, where stocks' tickers, spoken rationale, and affective signals (e.g., gestures, tone) may only align when considered together \cite{Gan2024MMEFinance, yanglet2025multimodalfinancialfoundationmodels}.

We address this gap by establishing baselines for financial SV understanding through topic-aligned captioning. Using the VideoConviction dataset \cite{galarnyk2025videoconviction}, we systematically test all single-, dual-, and tri-modality combinations of transcripts (T), audio (A), and video (V) across five tasks: main recommendation, sentiment analysis, video purpose, visual analysis, and financial entity recognition (Figure~\ref{fig:method}). Captions are evaluated for both fidelity (capturing correct stock ticker–action pairs) and accuracy, using G-VEval \cite{tong2024gvevalversatilemetricevaluating}. Results show consistent gains from multimodal integration over single-modality baselines, with selective subsets sometimes surpassing full tri-modal fusion \cite{mustakim2025watchlistenunderstandmislead}. These findings establish the first benchmarks for financial SV understanding and provide a foundation for analyzing domain-specific SV content.
\section{Related Work}
\label{sec:relatedwork}

Captioning has progressed from static image descriptions to temporally coherent video understanding \cite{ma2024imagecaptioningdynamicpath,chen2025vidcapbenchcomprehensivebenchmarkvideo}. Attention mechanisms \cite{vinyals2015show, xu2015show}, transformer architectures \cite{zhu2023topic}, and large-scale vision–language pretraining \cite{li2021clipit} have enabled topic-aligned and zero-shot captioning. However, most benchmarks emphasize generic or long-form domains, leaving short, domain-specific formats underexplored \cite{xiong2024lvd}.

In finance, early work focused almost exclusively on text, including earnings calls, regulatory filings, and market news \cite{FiQA, loughran2011liability, tatarinov2025languagemodelingfuturefinance, shah2025wordsuniteworldunified}. Sentiment analysis in particular often relied on lexicons such as Loughran–McDonald \cite{loughran2011liability}, before newer corpora captured finer-grained stance and subjectivity \cite{pardawala2024subjectiveqameasuringsubjectivityearnings, du-etal-2025-sentiment}. More recently, multimodal approaches have incorporated visual data such as financial charts \cite{shu2025finchartbenchbenchmarkingfinancialchart}, but short-form financial video remains unstudied. Prior work has not aligned finance-specific cues (e.g., stock tickers, charts, prosody, financial metrics) across modalities. We close this gap by establishing the first baselines for topic-aligned multimodal captioning of financial SVs, integrating transcripts, audio, and video in a unified evaluation.
\section{Experiments \& Data}
\label{sec:experiments}
\label{Sec:SVData}
We use the VideoConviction dataset \cite{galarnyk2025videoconviction}, which contains YouTube videos from financial influencers issuing explicit stock recommendations. Each full-length video (avg.~9 minutes) is manually segmented into shorter clips, each containing a single actionable recommendation of buy, sell, or hold. We filter the 687 recommendation clips for segments under 5 minutes to align with short-form videos \cite{gupta20223massiv}, yielding 624 SVs (avg.~4 minutes). Compared to ultra-short formats such as YouTube Shorts \cite{violot2024shorts}, video segments contain richer visual and auditory cues \cite{bernales2023effects}. Human annotators labeled the stock ticker and corresponding action for every segment, enabling quantitative evaluation.

The resulting segments provide three aligned modalities: \begin{enumerate}
\item \textbf{Video frames (V)} sampled at 0.25 fps following HourVideo \cite{chandrasegaran2024hourvideo1hourvideolanguageunderstanding}, consisting of various on-screen elements such as charts, overlays, and gestures.
\item \textbf{Audio (A)} preserving prosody and tonal cues.
\item \textbf{Transcript (T)} aligned to segment timestamps for precise mapping between speech and visuals.
\end{enumerate}
Together these modalities yield $2^3 - 1 = 7$ possible configurations: T, A, V, TA, TV, AV, and TAV.

\paragraph{Content Captioning}
\label{sec:content_captioning}
For each modality combination, we prompt MLLMs to generate concise, topic-aligned captions. This task requires chart interpretation, spatiotemporal reasoning, and multimodal grounding. Captions are evaluated across five topics: \textbf{Main Recommendation (MR)}, which captures the core financial advice with stock ticker, action, and rationale; \textbf{Sentiment Analysis (SA)}, which measures tone and confidence from multimodal cues; \textbf{Video Purpose (VP)}, which identifies communicative intent and target audience; \textbf{Visual Analysis (VIA)}, which describes charts, indicators, overlays, and gestures to test OCR and spatiotemporal reasoning; and \textbf{Financial Entities (FE)}, which extracts company names, tickers, and numerical values within the SV. Together these topics target core multimodal challenges, including visual grounding, prosody, domain-specific language, and entity recognition, providing a focused benchmark for MLLMs in financial video understanding \cite{ying2024mmtbenchcomprehensivemultimodalbenchmark, cho2025visionlanguagemodelsunderstand, bayoudh2022survey}.
\section{Evaluation}
\label{sec:evaluation}

We evaluate all seven modality combinations (T, A, V, TA, TV, AV, TAV) across five multimodal large language models (MLLMs): Qwen2.5-Omni-7B \cite{qwen_qwen2.5-omni-7b}, Gemini 2.0 Flash \cite{gemini2.0_flash}, Gemini 2.5 Flash \cite{gemini2.5_flash}, GPT-4o mini \cite{radford2024gpt4o}, and GPT-5 mini \cite{openai_gpt5mini_2025}. Since the GPT models do not support audio inputs, their evaluation is restricted to T, V, and TV combinations \cite{radford2024gpt4o, openai_gpt5mini_2025}

Performance is measured using two complementary metrics: (i) F1-score for ticker–action extraction, based on the stock ticker and action (buy, sell, and hold) labels annotated in the VideoConviction dataset \cite{galarnyk2025videoconviction}, and (ii) G-VEval \cite{tong2024gveval}, which scores accuracy on the remaining topics (MR, SA, VP, VIA, FE).

\subsection{Structured Stock Ticker + Action Extraction}
Table \ref{tab:f1_for_ticker_action} reports F1-scores across modality combinations. Gemini 2.0 Flash performs best with 0.60 under AV, while V alone is consistently weakest. T, A, and TA provide moderate gains, and performance peaks when V is fused with either T or A. This pattern highlights stock ticker–action extraction as an intrinsically multimodal task: text and audio supply the primary signal, while visuals add grounding through ticker placement, chart overlays, and on-screen prompts \cite{Rastogi_2020_CVPR_Workshops}. However, fusion is not always beneficial as TAV seldom achieves the best results \cite{Xia2023ModalityBias, Hori2017ABMF}.

\begin{table}[ht]
\centering
\begin{adjustbox}{width=\columnwidth}
\begin{tabular}{lccccccc}
\toprule
Model & T & A & V & TA & TV & AV & TAV \\
\midrule
Qwen 2.5
& 28 & 28 &\cellcolor{red!30}  9  & 32 & 29 & 34 &\cellcolor{green!30}  36 \\
Gemini 2.0 Flash 
& 46 & 55 &\cellcolor{red!30}  41 & 49 & 59 &\cellcolor{green!30}  \textbf{60} & 59 \\
Gemini 2.5 Flash  
& 38 & 41 &\cellcolor{red!30}  31 & 36 &\cellcolor{green!30}  50 & 46 & 46 \\
GPT-4o-mini
& 41 & -- &\cellcolor{red!30}  26 & -- & \cellcolor{green!30} 50 & -- & -- \\
GPT-5-mini 
& 40 & -- &\cellcolor{red!30}  32 & -- & \cellcolor{green!30} 57 & -- & -- \\
\bottomrule
\end{tabular}
\end{adjustbox}
\caption{F1-scores for stock ticker and action extraction across T, A, and V, with \colorbox{green!30}{best}, \colorbox{red!30}{worst}, and overall highest in \textbf{bold}.}
\label{tab:f1_for_ticker_action}
\end{table}

\subsection{Topic-aligned Captioning}

\paragraph{G-VEval Metric}

Evaluating captions in finance is challenging because small semantic shifts can change the meaning of a recommendation. Standard n-gram metrics (BLEU \citep{papineni2002bleu}, ROUGE \citep{lin2004rouge}, METEOR \citep{banerjee2005meteor}, CIDEr \citep{vedantam2015cider}) penalize valid paraphrases and often miss domain-specific nuances \citep{liu2024financial}. Hence, we use G-VEval, which leverages GPT-4o with chain-of-thought prompting to assess the accuracy of generated captions. Unlike CLIPScore \cite{hessel2021clipscore}, EMScore \cite{shi2021emscore}, and PAC-Score \cite{sarto2023positive}, which rely on zero-shot prompting and limited token windows, G-VEval \citep{tong2024gveval} uses structured reasoning to capture fine-grained, domain-specific semantics.

We organize our results by topic (MR, SA, VP, VIA, FE) to illustrate how different modality combinations contribute to each task. The following sections highlight the strengths and limitations of each model, revealing where each modality is most critical for short-form video understanding.

\paragraph{Main Recommendation (MR)}

\begin{table}[!h]
\centering
\setlength{\tabcolsep}{4pt}
\begin{tabular}{l
    >{\centering\arraybackslash}p{0.03\textwidth}
    >{\centering\arraybackslash}p{0.03\textwidth}
    >{\centering\arraybackslash}p{0.03\textwidth}
    >{\centering\arraybackslash}p{0.03\textwidth}
    >{\centering\arraybackslash}p{0.03\textwidth}
    >{\centering\arraybackslash}p{0.03\textwidth}
    >{\centering\arraybackslash}p{0.03\textwidth}}
\toprule
Model & T & A & V & TA & TV & AV & TAV \\
\midrule
Qwen 2.5           & 69 & 69 & \cellcolor{red!30}30 & 73 & 73 & 72 & \cellcolor{green!30}74 \\
Gemini 2.0 Flash   & 80 & 83 & \cellcolor{red!30}74 & 81 & 83 & \cellcolor{green!30} \textbf{85} & 84 \\
Gemini 2.5 Flash   & 73 & 71 & \cellcolor{red!30}45 & 71 & 75 & \cellcolor{green!30}80 & 76 \\
GPT-4o mini        & 78 & -- & \cellcolor{red!30}77 & -- & \cellcolor{green!30}82 & -- & -- \\
GPT-5 mini         & 62 & -- & \cellcolor{red!30}28 & -- & \cellcolor{green!30}76 & -- & -- \\
\bottomrule
\end{tabular}
\caption{G-VEval accuracy on topic \textbf{MR}, with \colorbox{green!30}{best}, \colorbox{red!30}{worst}, and overall highest in \textbf{bold}.}
\label{tab:MR}
\end{table}

MR, which captures the core financial advice, shows strong dependence on multiple modalities. As reported in Table \ref{tab:MR}, Gemini 2.0 Flash delivers the best overall performance, reaching an accuracy of 85 with the AV modality, while GPT-4o-mini follows closely with 82 under TV. These results underscore the importance of visual grounding—stock tickers, prices, and overlays—often cannot be recovered from transcripts alone. Qwen2.5 attains moderate results with TA and TV but fails with visual-only inputs, further suggesting that MR requires alignment between T, A, and V. Overall, the findings indicate that while text and audio contribute to capturing rationale and time frame, reliable extraction of the main recommendation depends critically on integrating multiple modalities.

\paragraph{Sentiment Analysis (SA)}

\begin{table}[ht]
\centering
\setlength{\tabcolsep}{4pt}
\begin{tabular}{l
    >{\centering\arraybackslash}p{0.03\textwidth}
    >{\centering\arraybackslash}p{0.03\textwidth}
    >{\centering\arraybackslash}p{0.03\textwidth}
    >{\centering\arraybackslash}p{0.03\textwidth}
    >{\centering\arraybackslash}p{0.03\textwidth}
    >{\centering\arraybackslash}p{0.03\textwidth}
    >{\centering\arraybackslash}p{0.03\textwidth}}
\toprule
Model & T & A & V & TA & TV & AV & TAV \\
\midrule
Qwen 2.5           & 61 & 55 & \cellcolor{red!30}48 & 57 & 66 & \cellcolor{green!30}70 & 68 \\
Gemini 2.0 Flash   & 70 & \cellcolor{red!30}69 & 70 & 70 & 74 & \cellcolor{green!30}75 & 73 \\
Gemini 2.5 Flash   & \cellcolor{red!30}73 & 74 & \cellcolor{green!30}79 & 74 & 77 & 76 & 76 \\
GPT-4o mini        & 69 & -- & \cellcolor{red!30}67 & -- & \cellcolor{green!30}71 & -- & -- \\
GPT-5 mini         & \cellcolor{red!30}71 & -- & \cellcolor{green!30}\textbf{82} & -- & 76 & -- & -- \\
\bottomrule
\end{tabular}
\caption{G-VEval accuracy on topic \textbf{SA}, with \colorbox{green!30}{best}, \colorbox{red!30}{worst}, and overall highest in \textbf{bold}.}
\label{tab:SA}
\end{table}

Table \ref{tab:SA} illustrates the difficulty of modeling tone and confidence in short-form financial videos. Visual input emerges as the strongest signal, consistent with prior findings that gestures, expressions, and delivery style are key for sentiment recognition \cite{Shafique2023NVC_CVPRW}. GPT-5 mini achieves the highest score (82) with video alone, underscoring the importance of visual cues. Gemini models show a similar vision-first pattern, though they maintain balanced performance across all modality combinations. In contrast, Qwen2.5 performs poorly with video alone (48) but improves substantially when audio is incorporated (70), suggesting reliance on audio cues. Overall, these results suggest that while sentiment is anchored in visual signals, complementary gains come from fusing audio and text, aligning with prior multimodal sentiment studies \cite{pereira2016fusingaudiotextualvisual,blanchard2018gettingsubtexttextscalable}.

\paragraph{Video Purpose (VP)}

\begin{table}[!h]
\centering
\setlength{\tabcolsep}{4pt}
\begin{tabular}{l
    >{\centering\arraybackslash}p{0.03\textwidth}
    >{\centering\arraybackslash}p{0.03\textwidth}
    >{\centering\arraybackslash}p{0.03\textwidth}
    >{\centering\arraybackslash}p{0.03\textwidth}
    >{\centering\arraybackslash}p{0.03\textwidth}
    >{\centering\arraybackslash}p{0.03\textwidth}
    >{\centering\arraybackslash}p{0.03\textwidth}}
\toprule
Model & T & A & V & TA & TV & AV & TAV \\
\midrule
Qwen 2.5           & 61 & \cellcolor{red!30}44 & 73 & 47 & \cellcolor{green!30}79 & 75 & 78 \\
Gemini 2.0 Flash   & \cellcolor{red!30}83 & 84 & \cellcolor{green!30}87 & 84 & \cellcolor{green!30}87 & \cellcolor{green!30}87 & \cellcolor{green!30}87 \\
Gemini 2.5 Flash   & \cellcolor{red!30}85 & 86 & \cellcolor{green!30}\textbf{89} & 86 & 88 & 87 & 87 \\
GPT-4o mini        & \cellcolor{red!30}78 & -- & \cellcolor{green!30}81 & -- & \cellcolor{green!30}81 & -- & -- \\
GPT-5 mini         & \cellcolor{red!30}74 & -- & \cellcolor{green!30}84 & -- & 79 & -- & -- \\
\bottomrule
\end{tabular}
\caption{G-VEval accuracy on topic \textbf{VP}, with \colorbox{green!30}{best}, \colorbox{red!30}{worst}, and overall highest in \textbf{bold}.}
\label{tab:VP}
\end{table}

VP targets the communicative intent and intended audience of a video. Table \ref{tab:VP} shows that performance is consistently vision-dominant, with Gemini 2.5 Flash reaching the highest accuracy (89) under the V modality. Scene composition, sponsor tags, and instructional overlays provide strong cues for purpose classification, and across models V alone often matches or exceeds multimodal fusion. These results suggest that VP is determined primarily by how content is packaged visually, with transcripts and audio contributing little additional signal.

\paragraph{Visual Analysis (VIA)}

\begin{table}[ht]
\centering
\setlength{\tabcolsep}{4pt}
\begin{tabular}{l
    >{\centering\arraybackslash}p{0.03\textwidth}
    >{\centering\arraybackslash}p{0.03\textwidth}
    >{\centering\arraybackslash}p{0.03\textwidth}
    >{\centering\arraybackslash}p{0.03\textwidth}
    >{\centering\arraybackslash}p{0.03\textwidth}
    >{\centering\arraybackslash}p{0.03\textwidth}
    >{\centering\arraybackslash}p{0.03\textwidth}}
\toprule
Model & T & A & V & TA & TV & AV & TAV \\
\midrule
Qwen 2.5           & 20 & \cellcolor{red!30}16 & \cellcolor{green!30}85 & 19 & 83 & 83 & 83 \\
Gemini 2.0 Flash   & \cellcolor{red!30}20 & 27 & \cellcolor{green!30}87 & 18 & \cellcolor{green!30}87 & \cellcolor{green!30}87 & \cellcolor{green!30}87 \\
Gemini 2.5 Flash   & 14 & \cellcolor{red!30}9 & \cellcolor{green!30}90 &\cellcolor{red!30} 9 & \cellcolor{green!30}90 & \cellcolor{green!30}90 & 89 \\
GPT-4o mini        & \cellcolor{red!30}51 & -- & 80 & -- & \cellcolor{green!30}81 & -- & -- \\
GPT-5 mini         & \cellcolor{red!30}24 & -- & \cellcolor{green!30}\textbf{92} & -- & 91 & -- & -- \\
\bottomrule
\end{tabular}
\caption{G-VEval accuracy on topic \textbf{VIA}, with \colorbox{green!30}{best}, \colorbox{red!30}{worst}, and overall highest in \textbf{bold}.}
\label{tab:VIA}
\end{table}

VIA evaluates OCR and spatiotemporal reasoning by requiring models to interpret charts, indicators, overlays, and gestures. Table \ref{tab:VIA} shows that performance is overwhelmingly vision-dominant. Accuracy collapses with text or audio alone (e.g., Gemini 2.5 Flash at 14 for T and 9 for A, GPT-5 mini at 24 for T) but rises to 90 and 92 with visual inputs alone for Gemini 2.5 Flash and GPT-5 mini, respectively. Adding transcripts or audio (TV, AV, TAV) offers no major benefit and sometimes reduces performance, indicating that non-visual modalities introduce redundancy or noise rather than complementary signal. These findings confirm VIA as a fundamentally video-driven task, where robust chart and overlay interpretation cannot be approximated through speech or text alone.

\paragraph{Financial Entities (FE)}

\begin{table}[h!]
\centering
\setlength{\tabcolsep}{4pt}
\begin{tabular}{l
    >{\centering\arraybackslash}p{0.03\textwidth}
    >{\centering\arraybackslash}p{0.03\textwidth}
    >{\centering\arraybackslash}p{0.03\textwidth}
    >{\centering\arraybackslash}p{0.03\textwidth}
    >{\centering\arraybackslash}p{0.03\textwidth}
    >{\centering\arraybackslash}p{0.03\textwidth}
    >{\centering\arraybackslash}p{0.03\textwidth}}
\toprule
Model & T & A & V & TA & TV & AV & TAV \\
\midrule
Qwen 2.5           & \cellcolor{red!30}65 & 68 & \cellcolor{green!30}79 & 66 & 78 & 78 & 78 \\
Gemini 2.0 Flash   & 78 & 78 & \cellcolor{green!30}85 & \cellcolor{red!30}77 & 83 & 83 & 82 \\
Gemini 2.5 Flash   & \cellcolor{red!30}76 & 80 & \cellcolor{green!30}86 & 80 & \cellcolor{green!30}86 & \cellcolor{green!30}86 & \cellcolor{green!30}86 \\
GPT-4o mini        & \cellcolor{red!30}76 & -- & \cellcolor{green!30}83 & -- & 81 & -- & -- \\
GPT-5 mini         & \cellcolor{red!30}78 & -- & \cellcolor{green!30} \textbf{87} & -- & \cellcolor{green!30} \textbf{87} & -- & -- \\
\bottomrule
\end{tabular}
\caption{G-VEval accuracy on topic \textbf{FE}, with \colorbox{green!30}{best}, \colorbox{red!30}{worst}, and overall highest in \textbf{bold}.}
 \label{tab:FE}
\end{table}

FE targets company names, stock tickers, and numerical values. Table \ref{tab:FE} shows a clear reliance on vision, with V yielding the strongest results across models and GPT-5 mini achieving peak accuracy of 87 under both V and TV. Unlike VIA, where visuals dominate due to charts and layouts, FE depends on structured identifiers that are typically displayed on-screen. Text and audio provide only partial cues and often add noise, offering little benefit over visual input alone. Overall, FE is a vision-dominant task, distinguished from VIA by its focus on precise finance-specific entities rather than broader visual context.
\section{Discussion}

Our findings reveal two consistent patterns. First, full tri-modal fusion does not guarantee superior performance over selective modality combinations \cite{zheng2025mllms}. In several cases, particularly VIA, FE, and VP, adding modalities to V introduced contradictory signals that reduced accuracy or induced hallucination \cite{zhang2025robustmultimodallargelanguage, lee2025optflowtextavs}. This suggests that careful modality selection, rather than indiscriminate fusion, is essential for robust multimodal reasoning. 

Second, the role of V varies sharply across task types. For explicitly visual tasks such as FE and VIA, models succeed at extracting precise financial details (e.g., “the stock price of \$49.39”). By contrast, for tasks where the financial signal is not directly visible, such as MR, visual input often defaults to scene-level descriptors (e.g., speaker appearance) instead of structured financial content. For SA and VP, contextual visual cues such as camera setup, gestures, and delivery style proved critical for interpreting tone and audience targeting. These trends point to a division of labor across modalities: vision should serve as the primary anchor, with audio and text acting as task-sensitive supplements. Effective prompting strategies are critical to ensure models attend to the right visual features.

Finally, our study is scoped to financial short videos drawn from the VideoConviction dataset. While this controlled setting enables systematic benchmarking, it also limits generalizability. Other short-form video domains may exhibit different multimodal dynamics, making cross-domain extensions a promising direction for future work.
\section{Conclusion}
\label{sec:conclusion}

We presented, to the best of our knowledge, the first benchmark for topic-aligned captioning of financial SVs. Our evaluation spans five tasks: MR, SA, and VP, which capture core financial advice, sentiment, and communicative intent; VIA, which probes visual grounding; and FE and ticker–action extraction, which test entity recognition and multimodal reasoning. Together, these tasks provide a structured foundation for assessing MLLMs in domains where dense jargon and layered visual cues make comprehension uniquely demanding \cite{zhang2025mmvidqa}.  

Our results demonstrate that selective modality fusion can outperform full tri-modal integration, underscoring the need for careful input design in multimodal reasoning \cite{gong2025temcoco, lee2025optflowtextavs}. Given the market influence of financial video recommendations, accurate topic-aligned captions are critical both for research and real-world applications. Looking ahead, we expect these benchmarks to guide future development of multimodal models in finance and to serve as a stepping stone for captioning in other high-stakes, domain-specific settings \cite{Kazakos2025GROVE}.

\paragraph{Acknowledgment} We appreciate the generous support and funding provided by the Georgia Institute of Technology Financial Services Innovation Lab and the Center for Finance and Technology.
{
    \small
    \bibliographystyle{ieeenat_fullname}
    \bibliography{main}
}

\end{document}